\documentclass[runningheads, envcountsame, a4paper]{llncs}
\usepackage[hyphens]{url}
\usepackage{graphics}
\usepackage{amsmath}
\usepackage[pdflatex]{graphicx}
\usepackage{epstopdf}
\usepackage{microtype}
\usepackage{xcolor}
\usepackage[misc]{ifsym}
\usepackage[ruled,vlined]{algorithm2e}

\begin{document}
\title{Diversity-Based Generalization for Unsupervised Text Classification under Domain Shift\thanks{To appear at 2020 European Conference on Machine Learning and Principles and Practice of Knowledge Discovery in Databases (ECML-PKDD).}}
\titlerunning{Diversity-Based Generalization}

\author{Jitin Krishnan (\Letter) \and Hemant Purohit \and Huzefa Rangwala}
\institute{George Mason University, Fairfax, VA, USA \\
\email{$\{jkrishn2, hpurohit, rangwala $\}@gmu.edu}} 
\tocauthor{Krishnan et al.}
\authorrunning \tocauthor
\toctitle{Diversity-Based Generalization for Unsupervised Text Classification under Domain Shift}

\maketitle

\begin{abstract}
Domain adaptation approaches seek to learn from a source domain and generalize it to an unseen target domain. At present, the state-of-the-art unsupervised domain adaptation approaches for subjective text classification problems leverage 
unlabeled target data along with labeled source data. 
In this paper, we propose a novel method for domain adaptation of \textit{single-task} text classification problems based on a simple but effective idea of diversity-based generalization that does not require unlabeled target data 
but still matches the state-of-the-art in performance. 
%
Diversity plays the role of promoting the model to better generalize and be indiscriminate towards domain shift by forcing the model not to rely on same features for prediction. 
We apply this concept on the most explainable component of neural networks, the attention layer. To generate sufficient diversity, we create a multi-head attention model and infuse a diversity constraint between the attention heads such that each head will learn differently.
%
We further expand upon our model by tri-training and designing a procedure with an additional diversity constraint between the attention heads of the tri-trained classifiers. 
Extensive evaluation using the standard benchmark dataset of Amazon reviews and a newly constructed dataset of Crisis events shows that our fully unsupervised method matches with the competing baselines that uses unlabeled target data.
Our results demonstrate that machine learning architectures that ensure sufficient diversity can generalize better; encouraging future research to design ubiquitously usable learning models without using unlabeled target data. 

\keywords{Text Classification  \and Unsupervised Domain Adaptation \and Natural Language Processing \and Neural Networks}
\end{abstract}

\section{Introduction}

In natural language processing, domain adaptation of sequence classification problems has several applications ranging from sentiment analysis \cite{blitzer2007biographies} to classifying social media posts during crisis events \cite{alam2018domain}. Knowledge learned from one domain, book reviews for instance, can be adapted to predict examples from a different domain such as reviews of electronics. Similarly, information about resource-need events learned from one natural disaster can be adapted to predict events from an ongoing crisis~\cite{purohit2018social}. With the publication of Amazon reviews dataset \cite{blitzer2007biographies} consisting of around 25 different domains, cross-domain sentiment analysis became a common way to evaluate machine learning models for domain adaptation in text.

The top performing models in this line of research largely remain dependent on unlabeled target data. Although unlabeled data from the target domain tends to help, it is imperative to realize the extent of performance gain that source data alone can bring. 
We consider that the ideal criterion for no supervision in domain adaptation is having zero knowledge about the target domain beforehand; even if it is unlabeled.  
Our work can be viewed either as a strong baseline for future unsupervised cross-domain research that utilizes unlabeled target data or as a new direction in fully unsupervised domain adaptation without using any target data at all; which is necessary for tasks such as relevancy prediction for actionable information filtering in domains such as natural disasters that require timely and efficient methods. 
We scope our work to the following setting: \textbf{a)} \textit{single-task transfer}, \textbf{b)} \textit{single source and target}, and \textbf{c)} \textit{without labeled and unlabeled target data available during training}. We compare and contrast our unsupervised methods to the existing counterparts. We do not consider any supervised or minimally supervised approaches in this work.

\textbf{Contributions:} \textbf{a)} We present a novel diversity-based generalization method using a multi-head attention model for domain adaption in unsupervised text classification tasks. \textbf{b)} To further improve the generalizability of our model and utilize additionally available unlabeled source data, we design a tri-training procedure with an additional diversity constraint between the attention heads of the tri-trained classifiers. \textbf{c)} Addressing the existing evaluation gap in component-level performance analysis, we show a systematic and incremental creation of our models by creating strong unsupervised baselines and improving upon existing work. 

\section{Related Work}
Early works on domain adaptation such as Structural Correspondence Learning \cite{blitzer2006domain} make use of unlabeled target data to find a joint representation by automatically inducing correspondences among features from different domains. The importance of a good feature representation was later formally analyzed with a generalization bound by Ben-David et al. \cite{ben2007analysis}. These studies realized the importance of finding commonality in features or pivots and minimizing the difference between the domains. Pan et al. \cite{pan} proposed a spectral feature alignment method to align domain-specific and domain-independent words into unified clusters via simultaneous co-clustering in a common latent space. Later, introduction of deep learning and neural networks helped remedy the problems of manual pivot selection and discrete feature representations. In order to learn better higher level representations, Stacked Denoising Autoencoders (SDA) \cite{sda} were introduced. Along with SDA, a more efficient version called marginalized SDA \cite{mSDA} with low computational cost and scalability has been utilized successfully in cross-domain tasks \cite{ganin2016domain,glorot2011domain}. Domain-Adversarial training of Neural Networks (DANN) \cite{ganin2014unsupervised} was proposed to effectively utilize unlabeled target data to create a classifier that is indiscriminate toward different domains. In their work a negative gradient (gradient reversal) from a domain classifier branch is back-propagated to promote the features at the lower layers of the network incapable of discriminating domains. DANN became an essential component in many works that followed. Recent works such as Adversarial Memory Network (AMN) \cite{amn} bring interpretability by using attention to capture the pivots. Along with attention, they effectively use gradient reversal to learn domain indiscriminate features. Hierarchical Attention Network (HATN) \cite{li2018hierarchical} expands upon AMN by first extracting pivots and then jointly training pivot and non-pivot networks. Interactive Attention Transfer Network (IATN) \cite{iatn}, another closely related work to AMN and HATN, showed the importance of attending `aspect' information. Another line of research, that approached domain adaption through innovation in training procedure, is tri-training \cite{saito2017asymmetric,zhou2005tri}. Tri-training utilizes three independently trained classifiers; of which one is trained only on unlabeled target data, pseudo-labeled by the other two. The final prediction is done by majority voting. Multi-task tri-training (MT-Tri) \cite{ruder2018strong}, on the other hand, introduced an orthogonality constraint between the two classifiers such that it can be trained jointly, reducing the compute time. This constraint is one of the inspirations for our work. Although diversity could be achieved through other means, we also focus on `orthogonality'. All of these recent works used unlabeled target data for training classifiers. Our goal is to show that similar performance is achievable without using any target data at all.

Based on how the dataset is used, approaches to domain adaptation can vary from minimally-supervised to unsupervised. Minimally-supervised approaches such as Aligned Recurrent Transfer \cite{collocate} utilize some \textit{labeled} data from the target domain, while unsupervised approaches such as DANN, AMN, HATN, or IATN utilize only \textit{unlabeled} target data making it a more realistic scenario in terms of usability where collecting labeled target data is expensive. However, many state-of-the-art unsupervised domain adaptation methods, strikingly, never compare with strong fully-unsupervised baselines where no target data is used. Newer methods have started using word vectors \cite{mikolov2013distributed} for their input word representations. However, the baselines they compare with, utilize large 5000-dimension feature vector of the most frequent unigrams and bigrams as the input representation. In addition, many recent works present a complex system without conducting a component-wise analysis which makes it unclear as to how much each component (word vectors, gradient reversal, or attention) contributed to the performance boost as compared to a simple DANN architecture. To address these evaluation gaps, we perform a systematic and incremental construction of architectures such that individual performance gain is realized. 

\textbf{Advantages and Practical Utility:} \textbf{a)} Our methods do not require any target data for training; making it out-of-the-box adaptable to any domain. \textbf{b)} We provide a method to utilize additionally available unlabeled source data. \textbf{c)} Our models are computationally cheaper (training converges quickly) when compared to the existing state-of-the-art models. \textbf{d)} Diversified attention can provide better quality of attended words which can be used for various downstream tasks such as knowledge graph construction.

\begin{figure}[h!]
  \centering 
    \includegraphics[width=12cm]{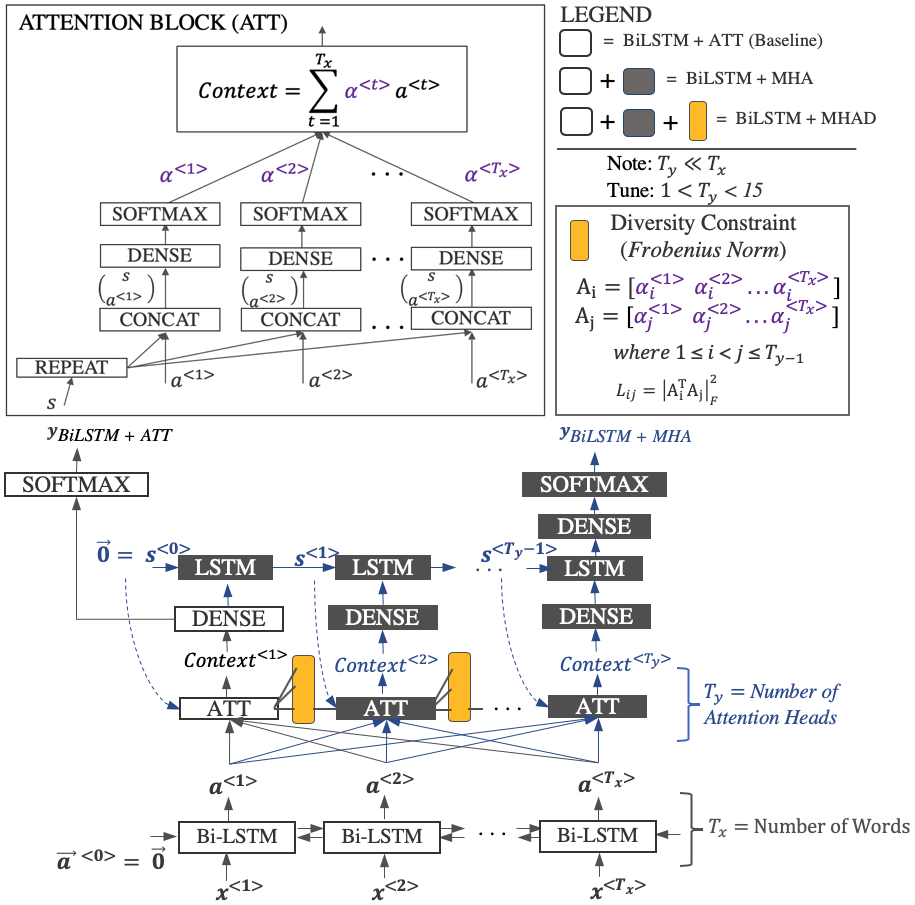}
 \caption{Complete architecture of the multi-head attention model with diversity.} \label{fig:bila}
\end{figure}

\section{Methodology}
\subsection{Problem Definition and Notations}

Given a source ($D_s$) and a target ($D_t$) domain, the goal is to train a classifier using data \textbf{only} from $D_s$ and predict examples from the completely unseen $D_t$. $X_{s}$ and $X_{t}$ represent the set of labeled data from source and target domains respectively with their corresponding ground truth labels $y_{s}$ and $y_{t}$. $X_{t}$ and $y_{t}$ are used for testing purposes only. $X_{s}^u$ and $X_{t}^u$ represent unlabeled data available from the source and target domain respectively. $X_{t}^u$ (used in all of our competing models either for adversarial training or tri-training) is \textbf{never} used in our models. Finally, $[.]^{pl}$ represents data that is pseudo-labeled by the classifier. To summarize:\\ \textit{\textbf{Input:}} $X_{s}$, $y_{s}$  (and $X_{s}^u$ for tri-training) \\
\textit{\textbf{Output:}} $y_{t}^{pred}$ $\gets$ $predict(X_{t})$

\subsection{Diversity-based Models}
\begin{figure}[h!]
  \centering
    \includegraphics[width=12cm]{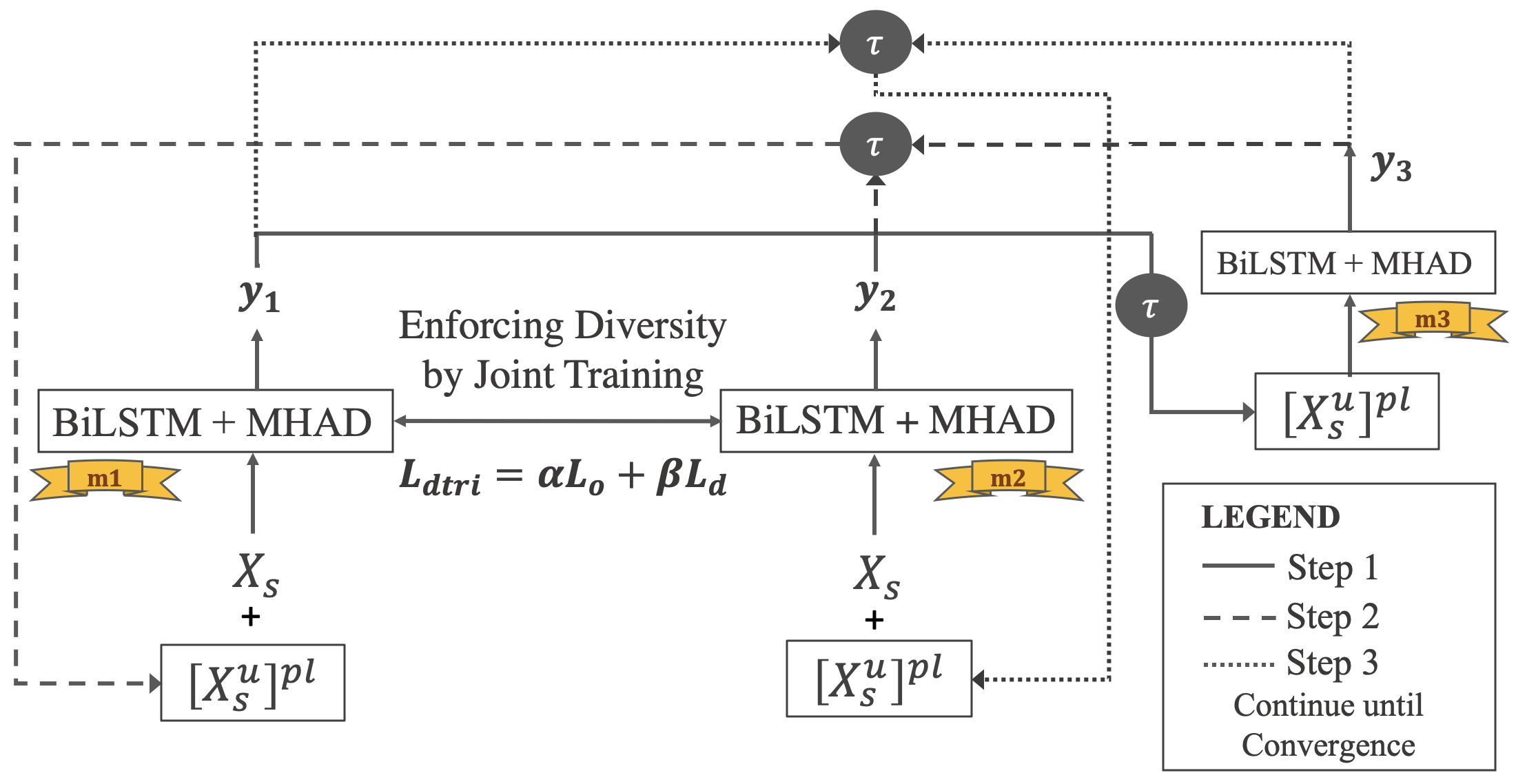}
 \caption{Tri-training BiLSTM+MHAD models} \label{fig:tri_training}
\end{figure}

We introduce 4 models with one integral concept: \textit{diversity}. Figure \ref{fig:bila} provides an overview of the first two models and Figure \ref{fig:tri_training} provides an overview of the last two. First is a multi-head attention baseline created to understand the naturally occurring diversity when multiple attention heads are connected. The second model enforces this diversity as a constraint such that all heads learn different features. The third model puts together three diversity-based classifiers and tri-trains them. Tri-training procedure in itself consists of an additional diversity constraint which forces two of the classifiers to learn differently. This is a one-step tri-training procedure intended for scenarios where no unlabeled source data is additionally available. When it is available, a full tri-training can be done until convergence, which is the fourth model.\\

\textbf{Multi-Head Attention for Sequence Classification (BiLSTM+MHA)}: BiLSTM+ATT is a standard baseline attention architecture constructed using BiLSTM \cite{hochreiter1997long,schuster1997bidirectional} and attention mechanism \cite{bahdanau2014neural,luong2015effective}. Bidirectional Long Short-Term Memory (BiLSTM) units have been successfully used in sequence modeling tasks because of their effectiveness in representing forward and backward dependencies in a sequence. For example, meanings of words like `good' and `bad' can be changed when they are prefixed with `not' or suffixed with `but'. Attention, on the other hand, provides task-specific benefits by attending the most relevant words such as `excellent' or `poor' in sentiment analysis. Attention and BiLSTM have been successfully combined previously for tasks such as relation extraction \cite{zhou2016attention} to capture important semantic information in a sentence.

BiLSTM+MHA is an extension of the BiLSTM+ATT baseline by adding multiple attention heads as shown in Figure \ref{fig:bila}. This is similar to machine-translation-like architecture \cite{luong2015effective} where each attention head leads to an LSTM cell with memory carried from previous cells to predict the next word. To customize it to classification purpose, we simply use the output from the final LSTM cell. Setting the classification task this way gives more leniency for the model to learn, remember, and generalize. Multiple attention heads can learn differently and what is learned from the previous heads is transferred to the next. However, this does not guarantee diversity as we do not know if the attention heads will in fact learn differently. In order to enforce diversity, we introduce the following models. \\

\textbf{Multi-Head Attention with Diversity (BiLSTM+MHAD)}: In order to guarantee that these attention heads learn differently and forcing the model not to rely on the same features, we create a \textit{diversity constraint}, an additional loss term shown below.
\begin{equation}
L_{d} = \frac{1}{k} \sum_{i=1}^{T_{y-2}} \sum_{j=i+1}^{T_{y-1}} \| A_{i}^{T} A_{j} \|^{2}_{F} \ ; \text{ where }  i\neq j 
\end{equation}

where $k=\frac{(T_{y}-2)(T_{y}-1)}{2}$, the total number of combinations. $T_{y}$ is the total number of attention heads. $A_i$ and $A_j$ are $i^{th}$ and $j^{th}$  attention heads and $\|.\|^{2}_{F}$ is the squared Frobenius norm, similar to the orthogonality constraint used in \cite{ruder2018strong}. We leave the last attention head from this loss term so that we have one layer that learns freely without any constraints. The complete architecture of this diversity-based model is shown in Figure \ref{fig:bila}. Resulting overall loss function, consisting of a binary cross entropy loss term and the diversity loss term, for $N$ training examples is shown below.
\begin{equation}
L(\theta) = -  \frac{1}{N} \sum_{i=1}^{N} [y_i \log \hat{y_i} +  (1-y_i) \log (1-\hat{y_i})] + \gamma L_{d} \label{eq:2}
\end{equation}
where $\gamma$ is the hyperparameter to control how much diversity to be enforced within the model.\\

\textbf{One-Step Diversity Tri-training (BiLSTM+MHAD-Tri-I)}: To further expand the concept of diversity, we tri-train the BiLSTM+MHAD models by adapting the multi-task tri-training procedure by \cite{ruder2018strong}. In addition to applying the diversity constraint within each classifiers, an additional orthogonality loss is enforced between first two models $m_1$ and $m_2$. The third model $m_3$ is left out from the joint training. The loss term is shown below.
\begin{equation}
L_{o} = \frac{1}{k} \sum_{i=1}^{T_{y}} \sum_{j=1}^{T_{y}} \| A(m_{1})_{i}^{T} A(m_{2})_{j} \|^{2}_{F} \
\end{equation}

where $k=\frac{(T_{y}-1)(T_{y})}{2}$. $A(m_{1})$ and $A(m_{2})$ are the attention heads for models $m_1$ and $m_2$ respectively. $T_{y}$ is the total number of attention heads of each model. The total tri-training diversity loss is given below.
\begin{equation}
L_{dtri} = \alpha L_{o} + \beta L_{d}  
\end{equation}
where $\alpha$ and $\beta$ are the hyperparameters to control how much diversity to be enforced within and between the models.

For one-step diversity tri-training shown in Algorithm \ref{alg:algorithm_onestep}, we jointly train $m_1$ and $m_2$ with tri-training diversity loss $L_{dtri}$. $m_3$ is separately trained as a BiLSTM+MHAD model. For predictions, a majority voting rule is applied over the three classifiers. The overall loss function for $N$ training examples is given below.
\begin{equation}
L(\theta) = -  \frac{1}{N} \sum_{i=1}^{N} [y_i \log \hat{y_i} +  (1-y_i) \log (1-\hat{y_i})] + L_{dtri} \label{eq:5}
\end{equation}\\

\begin{algorithm}[H]
\SetAlgoLined
 \textbf{Input}: $X_s$ \\
 \textbf{Output}: $m_1$, $m_2$, $m_3$ \\
 $m_1, m_2 \gets joint\_diversity\_train\_models(X_s)$ \\
 $m_3 \gets diversity\_train\_model(X_s)$ \\
 apply majority vote over $m_i$
 \caption{One-Step Diversity Tri-training} \label{alg:algorithm_onestep}
\end{algorithm}

\textbf{Tri-training until Convergence (BiLSTM + MHAD-Tri-II)}: Full tri-training, shown in Algorithm \ref{alg:tritraining} and Figure \ref{fig:tri_training}, utilizes additionally available unlabeled source data. While the first two classifiers $m_1$ and $m_2$ are jointly trained on labeled source data, the third classifier $m_3$ is solely dedicated to the training over unlabeled data that is pseudo-labeled by $m_1$ and $m_2$. Similar to \cite{ruder2018strong}, we define a threshold value $\tau$ such that at least one out of the two models should predict with probability greater than $\tau$ to be considered successfully pseudo-labeled. We set $\tau$ to be $0.7$. Starting with second iteration, $m_1$ is trained jointly with $m_2$ using a combination of labeled source data and unlabeled source data pseudo-labeled by $m_2$ and $m_3$. During joint-training, we give priority to the primary model by setting the loss weights accordingly. For example, while joint-training $m_1$ with $m_2$, losses for the models can be minimized in a $2:1$ ratio, giving priority to $m_1$. We continue this process until a convergence condition is met: $m_1 \approx m_2 \approx m_3$. \\

\begin{algorithm}[H]
\SetAlgoLined
\textbf{Input}: $X_s$, $X_s^u$\\
\textbf{Output}: $m_1$, $m_2$, $m_3$ \\
 \While{convergence condition is not met}{
  \For{$i \in {1..3}$}{
    $X^{pl} \gets \emptyset$ \\
  \For{$x \in X_s^u$}{
    \If{$p_i(x) = p_k(x)(j,k \neq i)$}{
        $X^{pl} \gets X^{pl} \cup \{(x,p_j(x))\}$
    }
  }
 \uIf{$i = 3$}{
    $m_3 \gets {\color{blue}diversity\_train(X^{pl})}$  \tcp*{Eq. \ref{eq:2}}
 }
 \uElseIf{$i = 1$}{
    $m_1 \gets {\color{blue}joint\_diversity\_train(X_s \cup X^{pl}, m_2)}$ \tcp*{Eq. \ref{eq:5}}
 }
 \Else{
  $m_2 \gets {\color{blue}joint\_diversity\_train(X_s \cup X^{pl}, m_1)}$ \tcp*{Eq. \ref{eq:5}}
  }
  }
 }
 apply majority vote over $m_i$ \\
 \caption{Tri-training \cite{ruder2018strong} - {\color{blue}Modified}} \label{alg:tritraining}
\end{algorithm}

\section{Experimental Evaluation}

\subsection{Benchmark Dataset: Amazon Reviews}
We use the standard benchmark Amazon reviews dataset\footnote{\url{http://www.cs.jhu.edu/~mdredze/datasets/sentiment/}} \cite{blitzer2007biographies} which is widely used for cross-domain sentiment analysis. We consider four domains: Books (B), Kitchen (K), DVD (D), and Electronics (E). For a fair evaluation of the architectures, we use the exact same raw dataset\footnote{\url{https://github.com/hsqmlzno1/HATN/tree/master/raw_data}} used by our top competitor model HATN \cite{li2018hierarchical}, which is a part of Blitzer's original raw dataset. We also use the same 300-dimensional word vectors\footnote{\url{https://code.google.com/archive/p/word2vec/}} \cite{mikolov2013distributed}. Table \ref{table:stat} summarizes this dataset.

\subsection{Crisis Dataset (Tweets)}

Additionally, we construct a new dataset consisting of Twitter posts (tweets) collected during three hurricane crises by \textit{CitizenHelper}~\cite{karuna2017citizenhelper} system: \textit{Harvey} and \textit{Irma} in 2017, and \textit{Florence} in 2018. Similar to sentiment classification, our goal here is to classify whether a tweet text indicates an event or not. Using the crowd-sourcing platform Figure-Eight\footnote{\url{https://www.figure-eight.com} now \url{https://appen.com/}}, three workers at minimum were assigned to give a binary label to each tweet. 
We define events to be actions that involve at least one noun/entity. Events could be past, present, or future actions. It could also be questions, news, or instructions about actions. Some examples are: `\textit{A rescues B}', `\textit{A is sending food to B}', `\textit{A will move to location B}', 
and so on. Table \ref{table:stat} summarizes this dataset. Unfortunately, the labeled dataset for Florence and Irma consists of very low number of positive events. Consequently, we set up the experiments such that we train only on \textit{Harvey} and test on \textit{Florence} and \textit{Irma}. 

\subsection{Experimental Setup}
We follow the traditional cross domain sentiment classification set up where each experiment consists of a source domain ($S$) and a target domain ($T$). A model will be trained on source data and tested on target data, represented as $S \rightarrow T$. We use all available labeled target data for testing. Crisis dataset is balanced before training and testing.

\begin{table}
\begin{center}
 \begin{tabular}{||p{1.3cm}| p{1.6cm} | p{1.6cm} | p{1.8cm} | p{2.95cm} | p{1.9cm}  ||} 
 \hline
  & \textbf{Positive} & \textbf{Negative} & \textbf{Unlabeled} & \textbf{Average Number of Tokens} & \textbf{Vocabulary}\\ [0.5ex] 
 \hline
 \textit{Books} & 3000 & 3000 & 9750 & 182.0 & 105920 \\ 
 \hline
 \textit{DVD} & 3000 & 3000  & 11843 & 197.5 & 117619  \\ 
 \hline
 \textit{Kitchen} & 3000  & 3000  & 13856 &102.0 & 52972   \\ 
 \hline
 \textit{Elec.} & 3000  & 3000  & 17009 &119.3 & 72458  \\ 
 \hline
 \hline
 \textit{Harvey} & 1122  & 960  & 10001 & 17.2 & 23562 \\
 \hline
 \textit{Florence} & 201  & 1475  & 10001 &17.1 & 26380  \\
 \hline
 \textit{Irma} & 313  & 596  & 10001 &15.3 & 20764 \\
 \hline
 \end{tabular}
\end{center}
\caption{Dataset Statistics}\label{table:stat}
\end{table}

\subsection{Implementation Details}

We use Keras deep learning library with Adam optimizer ($lr=0.005$, $beta_1=0.9$, $beta_2=0.999$, $decay=0.01$) for our implmentations. Maximum epoch is 40 with an early stopping patience of 3. Batch size is 32 and validation split is 0.15. 

We set the number of attention heads, $T_y = 5$ and number of words from each review, $T_x = 200$. To keep the model simple, we do not change this further. Dropouts are kept at $0.4$. $\tau$ is kept at $0.7$ and tri-training is stopped at $85\%$ agreement. We set $\gamma=0.01$, $\alpha=0.05$ and $\beta =0.01$. These values are obtained by performing a basic hyperparameter tuning using grid search.

\subsection{Baselines and Modifications}

\textbf{Adversarial Learning Based Methods:} DANN \cite{ganin2014unsupervised} introduced adversarial training by making use of unlabeled target domain data. Earlier layers of the deep neural network architecture are made domain invariant through back-propagating a negative gradient using a jointly trained domain classifier. It uses a 5000-dimensional feature vector of the most frequent unigrams and bigrams as the input representation. DAmSDA \cite{ganin2016domain}, on the other hand, uses mSDA \cite{mSDA} representation instead. We report the scores for DAmSDA and DANN from HATN \cite{li2018hierarchical}. For DANN, additionally, we create a customized implementation (\textbf{DANN$^+$}) using BiLSTM and word vectors. This modified architecture simply consists of a shared BiLSTM layer followed by a dense layer for sentiment classification and the same BiLSTM layer followed by a gradient reversal layer and a dense layer for domain classification. Note that the accuracy for \textbf{DANN$^+$} (our improved DANN) is $\textbf{+3.2\%}$ higher than what is reported in HATN.

\textbf{Tri-training Based Methods:} Multi-task tri-training (MT-Tri) \cite{ruder2018strong} conducts tri-training on a multilayer perceptron model with an orthogonality loss between the final layers to enforce diversity between the jointly trained models. Unlabeled target data pseudo-labeled by the first two classifiers are fed to the third classifier. Three classifiers are optimized until none of the models’ predictions change.  We improve upon this model (\textbf{MT-Tri$^+$}) by using word vectors and BiLSTM.

\begin{figure}[h!]
  \centering
\includegraphics[width=12.2cm]{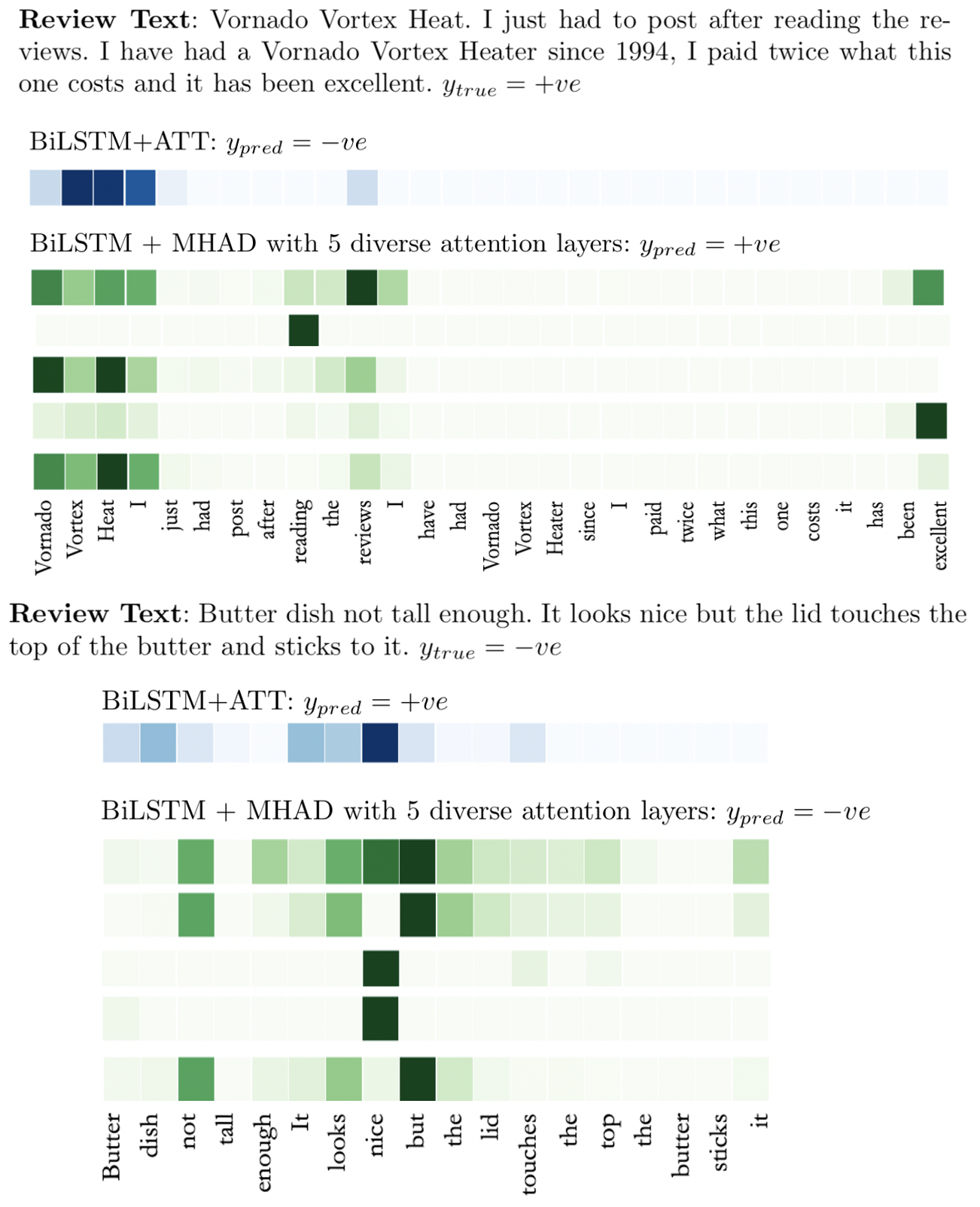}
 \caption{Two examples of \textit{kitchen} review predictions by BiLSTM + ATT and BiLSTM + MHAD models trained on \textit{book} reviews. When a single attention head fails to attend key words like `excellent' or `but', at least one of the diverse heads tends to make up for it.} \label{fig:example}
\end{figure}

\textbf{Attention Based Methods:} Recent works such as AMN \cite{amn}, HATN \cite{li2018hierarchical}, and IATN \cite{iatn} use attention to identify sentiment pivots. Utilizing unlabeled target data, gradient reversal is an essential component in their models for domain classification. AMN expands DANN to an attention-based model. HATN improves AMN further by building pivot and non-pivot networks. The pivot network (P-Net) performs the same task as AMN by extracting pivots. The non-pivot network (NP-Net) takes a transformed input that hides previously extracted pivots, which is then jointly trained with P-Net. IATN incorporates `aspect' information in addition to the sentence attentions. At the time of writing of this paper, the open source code\footnote{https://github.com/1146976048qq/IATN} for IATN is still being prepared by its authors. We include the reported scores for reference purpose. IATN reports a $0.8\%$ increase in performance as compared to HATN ($85.9\%$ versus $85.1\%$). IATN uses the same input settings and the dataset as HATN with one difference: 200-dimensional word vectors instead of 300. Meanwhile, we use the exact same dataset and GoogleNews word vectors used by HATN for all our experiments for both reproducibility as well as blind comparison.

\textbf{Strong Unsupervised Baselines:} To study component-wise performance, we construct two strong unsupervised baselines from standard neural network architectures: BiLSTM and BiLSTM+ATT. BiLSTM consists of traditional BiLSTM units with the final unit making the prediction. BiLSTM+ATT, as shown in Figure \ref{fig:bila}, adds a single attention layer on top of BiLSTM and the prediction is based on the output from the attention layer. Note that these two baselines still produce strong results and provide a reference for how much improvement following models make.

\begin{table}
\centering
 \caption{Classification accuracy scores showing that unlabeled target data is not necessary to achieve strong performance. \textbf{+}: improved implementations,  $\diamondsuit$: reproduced implementations, $\spadesuit$: strong unsupervised baselines constructed from standard neural network architectures, \textbf{*}: reported scores from \cite{li2018hierarchical,iatn} (see description). Our scores are averaged over 5 independent runs. \textbf{Note that only the models in the bottom table \underline{do not} use any unlabeled target data.}} \label{table:result}
\scalebox{0.9}{
\begin{tabular}{|c||cccccp{1.1cm}c| }
    \hline
 S $\rightarrow$ T & DANN* & DAmSDA* & IATN* & DANN$^+$ & MT-Tri$^+$ & AMN$^\diamondsuit$ /P-Net & HATN$^\diamondsuit$ \\ 
 \hline
B $\rightarrow$ D & 83.42  &  86.12  &  86.80  &  82.85  &  84.67  &  87.07  &  87.70   \\
B $\rightarrow$ E &76.27  &  79.02  &  86.50  &  81.03  &  84.62  &  82.98  &  86.20    \\
B $\rightarrow$ K & 77.90 &  81.05  &  85.90  &  82.01  &  84.78  &  84.85  &  87.08    \\
K $\rightarrow$ B & 74.17  &  80.55  &  84.70  &  79.38  &  80.98  &  83.50  &  84.83    \\
K $\rightarrow$ D & 75.32  &  82.18  &  84.40  &  79.04  &  78.89  &  82.83  &  84.73    \\
K $\rightarrow$ E & 85.53  &  88.00  &  87.60  &  86.00  &  85.87  &  86.72  &  89.08    \\
E $\rightarrow$ B & 73.53  &  79.92  &  81.80  &  78.92  &  80.64  &  83.28  &  83.62    \\
E $\rightarrow$ K & 84.53  &  85.80  &  88.70  &  86.43  &  89.62  &  89.80  &  90.12    \\
E $\rightarrow$ D &76.27  &  82.63  &  84.10  &  77.83  &  79.97  &  83.37  &  83.87    \\
D $\rightarrow$ B & 80.77  &  85.17  &  87.00  &  84.32  &  85.67  &  87.85  &  88.02   \\
D $\rightarrow$ E & 76.35  &  76.17  &  86.90  &  81.74  &  84.48  &  84.65  &  86.78    \\
D $\rightarrow$ K & 78.15  &  82.60  &  85.80  &  83.29  &  85.05  &  84.28  &  87.00    \\
\hline
AVG & 78.52  &  82.43  &  85.90  &  81.78  &  83.77  &  85.10  &  \textbf{86.59}\\
\hline
\end{tabular}
}

\scalebox{0.9}{
\begin{tabular}{|c||p{1.6cm}p{1.6cm}p{1.6cm}p{1.6cm}p{2.3cm}p{2.2cm}| }
    \hline
 S $\rightarrow$ T & BiLSTM$^\spadesuit$ & BiLSTM +ATT$^\spadesuit$ &BiLSTM +MHA & BiLSTM +MHAD & BiLSTM +MHAD-Tri-I & BiLSTM +MHAD-Tri-II  \\ 
 \hline
B $\rightarrow$ D &  84.19  &  87.44 & 87.29  &  87.54  &  87.76  &  87.46  \\
B $\rightarrow$ E &  83.61  &  83.90 & 85.36  &  85.63  &  85.75  &  86.08  \\
B $\rightarrow$ K &  83.87  &  85.21 & 86.04  &  87.06  &  87.34  &  87.68  \\
K $\rightarrow$ B &  80.52  &  82.15 & 83.11  &  83.70  &  84.19  &  84.23  \\
K $\rightarrow$ D &  78.28  &  80.17 & 81.50  &  82.27  &  82.11  &  83.34  \\
K $\rightarrow$ E &  86.33  &  87.30 & 88.60  &  88.81  &  88.98  &  89.22  \\
E $\rightarrow$ B &  80.58  &  82.10 & 83.55  &  83.67  &  83.96  &  84.33  \\
E $\rightarrow$ K &  88.07  &  88.19 & 89.61  &  89.96  &  90.07  &  91.05  \\
E $\rightarrow$ D &  78.08  &  81.93 & 82.77  &  82.93  &  82.87  &  82.81  \\
D $\rightarrow$ B &  83.93  &  87.72 & 87.77  &  88.22  &  88.51  &  88.74  \\
D $\rightarrow$ E &  82.98  &  84.57 & 84.75  &  85.93  &  85.79  &  86.21  \\
D $\rightarrow$ K &  84.38  &  85.45& 86.50  &  86.73  &  86.74  &  87.37  \\
\hline
AVG &  82.90  &  84.68 & 85.57  &  85.98  &  86.17  &  \textbf{86.54} \\
\hline
\end{tabular}
}

\end{table}


\begin{table}
\centering
\caption{Classification accuracy scores for crisis dataset.} \label{table:harvey_result}
\scalebox{0.9}{
\begin{tabular}{|c||p{1.1cm} p{1.4cm} p{1.4cm} p{1.4cm} p{2.2cm} p{2.2cm}| }
\hline
 S $\rightarrow$ T & HATN & BiLSTM +ATT & BiLSTM +MHA & BiLSTM +MHAD & BiLSTM +MHAD-Tri-I & BiLSTM +MHAD-Tri-II \\
 \hline
 H $\rightarrow$ F & 80.01 & 74.88 & 74.32 & 75.69 & 76.00 & 78.11 \\
 H $\rightarrow$ I & 58.53 & 63.84 & 64.32 & 65.10 & 65.02 & 64.38 \\
\hline
\end{tabular}
}
\end{table}

\begin{table}
\centering
 \caption{
 Training time in d-hh:mm:ss for H$\rightarrow$F on a Dual Intel(R) Xeon(R) Gold 5120 CPU@2.2GHz with 28 cores and 1.5TB RAM.
} \label{table:runtime}
\scalebox{0.9}{
\begin{tabular}{|c||c|c|c|c| }
\hline
HATN & BiLSTM+MHA & BiLSTM+MHAD & BiLSTM+MHAD-Tri-I & BiLSTM+MHAD-Tri-II \\
 \hline
 1-08:31:09 & 00:50:31 & 01:29:28 & 2:07:23 & 6:21:18 \\
\hline
\end{tabular}
}
\end{table}

\section{Results \& Discussion}

Tables \ref{table:result}, \ref{table:harvey_result}, and \ref{table:runtime} show the competitive nature of our fully unsupervised methods when compared with the existing unsupervised counterparts that use unlabeled target data. Our experiments showed incrementally improving results when each component is added to the baselines. Attention with diversity improved the single-attention baseline and tri-training with diversity improved it even further. Using additionally available unlabeled source data proved to be fruitful for most of the domains. Note that for crisis dataset we only use \textit{Harvey} for training because labeled data for \textit{Florence} and \textit{Irma} was just too low.

An implication of the diversity-based attention heads is shown in Figure \ref{fig:example}. Diversity pushes the model not to rely on the same features. First example shows misclassification by a single attention model that attends incorrect sentiment words like `Vortex' and `Heat'. However, with diversity, the model is lenient and look for alternate features. At least one of the $T_y$ diverse heads tends to find important words like `excellent'. These examples also show that placing diversity on attention layers, rather than on any other hidden layers, provides an explainable understanding of which words the model deems to be important and can be used for subsequent pivot extraction like in AMN or  HATN.\\

\textbf{Computational Performance:} To show that our work is practically useful for all communities alike, experiments are run on a CPU. A sample training time comparison is shown in Table \ref{table:runtime}. HATN needs gradient reversal to utilize unlabeled target data for the domain classifier branch and pivot extraction for joint training; subsequently making it much slower.\\

\textbf{Gradient Reversal:} To study the impact of gradient reversal procedure with BiLSTM
, we conducted experiments with unlabeled target data. The performance of BiLSTM versus DANN$^+$ (improved DANN) models in Table \ref{table:result} showed that, with a good dropout value for the BiLSTM units, gradient reversal did not help much. On a similar note, domain adversarial loss was found not to be helpful in tri-training experiments \cite{ruder2018strong}. In our context, we speculate that this might be because the dropout in the BiLSTM layer drops individual words that can lead to a better generalization which is essentially the purpose of gradient reversal. 
This will be studied in our future work.

\begin{table}
\centering
\caption{Classification accuracy scores for three distinct combinations.} \label{table:addi}
\scalebox{0.9}{
\begin{tabular}{|p{3.0cm}||p{1.0cm}|p{2.2cm}| }
\hline
 S $\rightarrow$ T & P-Net & BiLSTM +MHAD-Tri-II \\
 \hline
 \textit{Electronics} $\rightarrow$ \textit{Yelp} & 88.45 & 89.15 \\
\textit{Kitchen} $\rightarrow$ \textit{IMDb} & 76.38 & 78.33  \\
\textit{Yelp} $\rightarrow$ \textit{IMDb} & 78.75 & 77.28 \\
\hline
\end{tabular}
}
\end{table}

\textbf{Additional Analysis:} Generalizability of our models is further tested with three randomly selected experiments using very divergent domains such as Yelp\footnote{https://www.yelp.com/dataset/challenge} restaurant reviews and IMDb \cite{maas2011learning} movie reviews in addition to Amazon reviews; \textit{Electronics (Amazon)} $\rightarrow$ \textit{Yelp}, \textit{Kitchen (Amazon)} $\rightarrow$ \textit{IMDb}, and \textit{Yelp} $\rightarrow$ \textit{IMDb}. We randomly selected $2000$ positive and negative reviews from Yelp and IMDb. Their accuracy scores on our final model when compared to PNet\footnote{PNet is the first component of HATN which is computationally faster and within $\sim1.5\%$ accuracy of HATN} of HATN is shown in Table \ref{table:addi}. Once again, this shows that unlabeled target data is not always necessary; thus providing us with a fully unsupervised and computationally efficient alternative for domain adaptation in text classification tasks.

\section{Future Work}
Experiments shown in the additional analysis section can be expanded to a lot more datasets and divergent domains; particularly applying to domains where unlabeled target data is not readily available such as during an onset of a natural or man-made disaster event. A careful empirical study of several existing complex architectures that employ adversarial training using the gradient reversal strategy is another direction. It is crucial to understand how much of the gain claimed by the adversarial training approach can actually be brought through by good generalization of the model without using any unlabeled data from the target domain. Recent progress in deep learning and natural language processing has seen impressive performance gain across various domains by using the transformer \cite{devlin2018bert,vaswani2017attention} based models. Leveraging such models and studying the effect of encouraging diversity among the architectures appear to be a promising future direction. 

\section{Conclusion}
Our study shows that machine learning architectures designed for achieving sufficient diversity in learning can generalize better for domain adaptation. Further, unlabeled target data, used often by state-of-the-art models, is not always necessary to produce strong performance for domain adaptation in subjective text classification problems. We introduced a novel diversity-based generalization approach for the domain shift problem using a multi-head attention model where attention heads are constrained to learn differently such that the classifier can leverage on alternative features. 
Experiments on the standard benchmark dataset of Amazon reviews and a newly constructed dataset of Crisis events showed that our fully unsupervised methods that completely avoid target data can indeed match the competing unsupervised baselines.\\ 

\noindent \textbf{Reproducibility:} Code, datasets, and documentation are available at -\\ 
{\color{blue}\url{https://github.com/jitinkrishnan/Diversity-Based-Generalization}}

\section*{Acknowledgements}
Authors would like to thank U.S. National Science Foundation grant IIS-1815459 for partially supporting this research.

\bibliographystyle{splncs04}
\bibliography{splncs04}

\begin{thebibliography}{10}
\providecommand{\url}[1]{\texttt{#1}}
\providecommand{\urlprefix}{URL }
\providecommand{\doi}[1]{https://doi.org/#1}

\bibitem{alam2018domain}
Alam, F., Joty, S., Imran, M.: Domain adaptation with adversarial training and
  graph embeddings. arXiv preprint arXiv:1805.05151  (2018)

\bibitem{bahdanau2014neural}
Bahdanau, D., Cho, K., Bengio, Y.: Neural machine translation by jointly
  learning to align and translate. arXiv preprint arXiv:1409.0473  (2014)

\bibitem{ben2007analysis}
Ben-David, S., Blitzer, J., Crammer, K., Pereira, F.: Analysis of
  representations for domain adaptation. In: Advances in neural information
  processing systems. pp. 137--144 (2007)

\bibitem{blitzer2007biographies}
Blitzer, J., Dredze, M., Pereira, F.: Biographies, bollywood, boom-boxes and
  blenders: Domain adaptation for sentiment classification. In: Proceedings of
  the 45th annual meeting of the association of computational linguistics. pp.
  440--447 (2007)

\bibitem{blitzer2006domain}
Blitzer, J., McDonald, R., Pereira, F.: Domain adaptation with structural
  correspondence learning. In: Proceedings of the 2006 conference on empirical
  methods in natural language processing. pp. 120--128 (2006)

\bibitem{mSDA}
Chen, M., Xu, Z., Weinberger, K., Sha, F.: Marginalized denoising autoencoders
  for domain adaptation. arXiv preprint arXiv:1206.4683  (2012)

\bibitem{collocate}
Cui, W., Zheng, G., Shen, Z., Jiang, S., Wang, W.: Transfer learning for
  sequences via learning to collocate. arXiv preprint arXiv:1902.09092  (2019)

\bibitem{devlin2018bert}
Devlin, J., Chang, M.W., Lee, K., Toutanova, K.: Bert: Pre-training of deep
  bidirectional transformers for language understanding. arXiv preprint
  arXiv:1810.04805  (2018)

\bibitem{ganin2014unsupervised}
Ganin, Y., Lempitsky, V.: Unsupervised domain adaptation by backpropagation.
  arXiv preprint arXiv:1409.7495  (2014)

\bibitem{ganin2016domain}
Ganin, Y., Ustinova, E., Ajakan, H., Germain, P., Larochelle, H., Laviolette,
  F., Marchand, M., Lempitsky, V.: Domain-adversarial training of neural
  networks. The Journal of Machine Learning Research  \textbf{17}(1),
  2096--2030 (2016)

\bibitem{glorot2011domain}
Glorot, X., Bordes, A., Bengio, Y.: Domain adaptation for large-scale sentiment
  classification: A deep learning approach. In: Proceedings of the 28th
  international conference on machine learning (ICML-11). pp. 513--520 (2011)

\bibitem{hochreiter1997long}
Hochreiter, S., Schmidhuber, J.: Long short-term memory. Neural computation
  \textbf{9}(8),  1735--1780 (1997)

\bibitem{karuna2017citizenhelper}
Karuna, P., Rana, M., Purohit, H.: Citizenhelper: A streaming analytics system
  to mine citizen and web data for humanitarian organizations. In: Eleventh
  International AAAI Conference on Web and Social Media. pp. 729--730 (2017)

\bibitem{li2018hierarchical}
Li, Z., Wei, Y., Zhang, Y., Yang, Q.: Hierarchical attention transfer network
  for cross-domain sentiment classification. In: Thirty-Second AAAI Conference
  on Artificial Intelligence (2018)

\bibitem{amn}
Li, Z., Zhang, Y., Wei, Y., Wu, Y., Yang, Q.: End-to-end adversarial memory
  network for cross-domain sentiment classification. In: IJCAI. pp. 2237--2243
  (2017)

\bibitem{luong2015effective}
Luong, M.T., Pham, H., Manning, C.D.: Effective approaches to attention-based
  neural machine translation. arXiv preprint arXiv:1508.04025  (2015)

\bibitem{maas2011learning}
Maas, A.L., Daly, R.E., Pham, P.T., Huang, D., Ng, A.Y., Potts, C.: Learning
  word vectors for sentiment analysis. In: The 49th Annual Meeting of the
  Association for Computational Linguistics (2011)

\bibitem{mikolov2013distributed}
Mikolov, T., Sutskever, I., Chen, K., Corrado, G.S., Dean, J.: Distributed
  representations of words and phrases and their compositionality. In: Advances
  in neural information processing systems. pp. 3111--3119 (2013)

\bibitem{pan}
Pan, S.J., Ni, X., Sun, J.T., Yang, Q., Chen, Z.: Cross-domain sentiment
  classification via spectral feature alignment. In: Proceedings of the 19th
  international conference on World wide web. pp. 751--760. ACM (2010)

\bibitem{purohit2018social}
Purohit, H., Castillo, C., Imran, M., Pandey, R.: Social-eoc: Serviceability
  model to rank social media requests for emergency operation centers. In: 2018
  IEEE/ACM International Conference on Advances in Social Networks Analysis and
  Mining (ASONAM). pp. 119--126. IEEE (2018)

\bibitem{ruder2018strong}
Ruder, S., Plank, B.: Strong baselines for neural semi-supervised learning
  under domain shift. arXiv preprint arXiv:1804.09530  (2018)

\bibitem{saito2017asymmetric}
Saito, K., Ushiku, Y., Harada, T.: Asymmetric tri-training for unsupervised
  domain adaptation. In: Proceedings of the 34th International Conference on
  Machine Learning-Volume 70. pp. 2988--2997. JMLR. org (2017)

\bibitem{schuster1997bidirectional}
Schuster, M., Paliwal, K.K.: Bidirectional recurrent neural networks. IEEE
  Transactions on Signal Processing  \textbf{45}(11),  2673--2681 (1997)

\bibitem{vaswani2017attention}
Vaswani, A., Shazeer, N., Parmar, N., Uszkoreit, J., Jones, L., Gomez, A.N.,
  Kaiser, {\L}., Polosukhin, I.: Attention is all you need. In: Advances in
  neural information processing systems. pp. 5998--6008 (2017)

\bibitem{sda}
Vincent, P., Larochelle, H., Lajoie, I., Bengio, Y., Manzagol, P.A.: Stacked
  denoising autoencoders: Learning useful representations in a deep network
  with a local denoising criterion. Journal of machine learning research
  \textbf{11}(Dec),  3371--3408 (2010)

\bibitem{iatn}
Zhang, K., Zhang, H., Liu, Q., Zhao, H., Zhu, H., Chen, E.: Interactive
  attention transfer network for cross-domain sentiment classification  (2019)

\bibitem{zhou2016attention}
Zhou, P., Shi, W., Tian, J., Qi, Z., Li, B., Hao, H., Xu, B.: Attention-based
  bidirectional long short-term memory networks for relation classification.
  In: Proceedings of the 54th Annual Meeting of the Association for
  Computational Linguistics (Volume 2: Short Papers). pp. 207--212 (2016)

\bibitem{zhou2005tri}
Zhou, Z.H., Li, M.: Tri-training: Exploiting unlabeled data using three
  classifiers. IEEE Transactions on Knowledge \& Data Engineering (11),
  1529--1541 (2005)

\end{thebibliography}

\end{document}